\documentclass{article}

\usepackage{algorithm}
\usepackage{amsfonts}
\usepackage{amsmath}
\usepackage{amssymb}
\usepackage{amsthm}
\usepackage{bm}
\usepackage{booktabs}
\usepackage{caption}
\usepackage{color}
\usepackage{graphicx}
\usepackage{hyperref}
\usepackage{microtype}
\usepackage{nicefrac}
\usepackage{siunitx}
\usepackage{stackengine}
\usepackage{subcaption}
\usepackage{url}
\usepackage{verbatim}
\usepackage{wrapfig}
\usepackage[accepted]{icml2018}

\newcommand{\E}{\mathbb{E}}

\definecolor{ta-color}{rgb}{0.1, 0.45, 0.9}
\definecolor{peb-color}{rgb}{0.1, 0.6, 0.1}

\icmltitlerunning{Time Limits in Reinforcement Learning}

\begin{document}

\twocolumn[
\icmltitle{Time Limits in Reinforcement Learning}

\begin{icmlauthorlist}
\icmlauthor{Fabio Pardo}{icl}
\icmlauthor{Arash Tavakoli}{icl}
\icmlauthor{Vitaly Levdik}{icl}
\icmlauthor{Petar Kormushev}{icl}
\end{icmlauthorlist}

\icmlaffiliation{icl}{Robot Intelligence Lab, Imperial College London, UK}

\icmlcorrespondingauthor{Fabio Pardo, Arash Tavakoli, Vitaly \mbox{Levdik}, Petar Kormushev}{f.pardo, a.tavakoli, v.levdik, p.kormushev \mbox{@imperial.ac.uk}}

\icmlkeywords{Reinforcement Learning, Deep Learning, Machine Learning, Control, Markov Decision Processes, ICML}

\vskip 0.3in
]

\printAffiliationsAndNotice{}

\begin{abstract}
In reinforcement learning, it is common to let an agent interact for a fixed amount of time with its environment before resetting it and repeating the process in a series of episodes. The task that the agent has to learn can either be to maximize its performance over (i) that fixed period, or (ii) an indefinite period where time limits are only used during training to diversify experience. In this paper, we provide a formal account for how time limits could effectively be handled in each of the two cases and explain why not doing so can cause state aliasing and invalidation of experience replay, leading to suboptimal policies and training instability. In case (i), we argue that the terminations due to time limits are in fact part of the environment, and thus a notion of the remaining time should be included as part of the agent's input to avoid violation of the Markov property. In case (ii), the time limits are not part of the environment and are only used to facilitate learning. We argue that this insight should be incorporated by bootstrapping from the value of the state at the end of each partial episode. For both cases, we illustrate empirically the significance of our considerations in improving the performance and stability of existing reinforcement learning algorithms, showing state-of-the-art results on several control tasks.
\end{abstract}

\section{Introduction}
\label{sec:intro}

The reinforcement learning framework \citep{Kaelbling:1996RLsurvey, Sutton:1998, Szepesvari:2010algorithms} considers a sequential interaction between an agent and its environment. At every time step $t$, the agent receives a representation $S_t$ of the environment's state, selects an action $A_t$ that is executed in the environment which in turn provides a representation $S_{t+1}$ of the successor state and a \mbox{reward signal $R_{t+1}$.} \mbox{An individual} reward received by the agent does not directly indicate the quality of its latest action as some rewards may be the consequence of a series of actions taken in the past. Thus, the goal of the agent is to maximize the discounted sum of future rewards, also known as the \textit{return}:
\begin{equation}
\label{equ:time-unlimited-return}
G_t = R_{t+1} + \gamma R_{t+2} + \gamma^2 R_{t+3} + ... = \sum_{k=1}^\infty {\gamma^{k-1} R_{t+k}}
\end{equation}
A discount factor $0 \leq \gamma < 1$ is necessary to exponentially decay the future rewards ensuring bounded returns. While the series is infinite, it is common to use this expression even in the case of possible terminations, such as timeouts, by considering them to be the entering of an absorbing state that transitions only to itself and generates zero rewards thereafter. However, when the maximum length of an episode is fixed, it is easier to rewrite the expression above by explicitly including the time limit $T$: 
\begin{equation}
\label{equ:time-limited-return}
G_{t:T} = R_{t+1} + ... + \gamma^{T-t-1} R_T = \sum_{k=1}^{T-t} {\gamma^{k-1} R_{t+k}}
\end{equation}
Optimizing for the expectation of the return specified in Equation~\ref{equ:time-limited-return} is suitable for naturally \textit{time-limited tasks} where the agent has to maximize its expected return only over a fixed episode length. In this case, since the return is bounded, a discount factor of $\gamma = 1$ can be used. However, in practice it is still common to keep $\gamma$ smaller than $1$ in order to give more priority to short-term rewards. Under this optimality model, the objective of the agent does not go beyond the time limit. Therefore, such an agent could for example learn to take more risky actions leading to higher expected returns as approaching the time limit. In Section~\ref{sec:time-aware}, we study this case and illustrate that due to the time limit terminations, the remaining time is an inherent part of the environment's state and is essential to its \textit{Markov property} \citep{Sutton:1998}. Therefore, we argue for the inclusion of a notion of the remaining time in the agent's input, an approach that we refer to as \textit{time-awareness} (TA). We describe various scenarios where lacking a notion of the remaining time can lead to suboptimal policies and instability, and demonstrate significant performance improvements for agents with time-awareness.

\begin{figure*}[t]
    \begin{subfigure}{0.24\textwidth}
        \centering
        \includegraphics[width=0.7\textwidth]{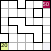}
        \caption{Gridworld}
        \label{subfig:gridworld}
    \end{subfigure}
    \hfill
    \begin{subfigure}{0.24\textwidth}
        \centering
        \includegraphics[width=0.7\textwidth]{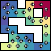}
        \caption{Standard}
        \label{subfig:gridworld_std}
    \end{subfigure}
    \hfill
    \begin{subfigure}{0.24\textwidth}
        \centering
        \includegraphics[width=0.7\textwidth]{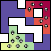}
        \caption{Time-awareness}
        \label{subfig:gridworld_ta}
    \end{subfigure}
    \hfill
    \begin{subfigure}{0.24\textwidth}
        \centering
        \includegraphics[width=0.7\textwidth]{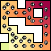}
        \caption{Partial-episode bootstrapping} \label{subfig:gridworld_peb}
    \end{subfigure}
    \caption{Color-coded state-values and policies learned by tabular Q-learning on our Two-Goal Gridworld task with $T=3$ and $-1$ penalties for movements. (a) The gridworld and its two goals. (b) The standard agent perceives timeout terminations as environmental ones and is not time-aware. It always tries to go for the closest goal even if the remaining time is not sufficient. (c) The time-aware agent maximizes its return over the finite horizon and learns to stay in place when there is not enough time to reach a goal. (d) The agent with partial-episode bootstrapping maximizes its return over an indefinite horizon, it learns to go for the most rewarding goal.}
    \label{fig:gridworld}
\end{figure*}

On the other hand, optimizing for the expectation of the return specified by Equation~\ref{equ:time-unlimited-return} is relevant for \textit{time-unlimited tasks} where the interaction is not limited in time by nature. In this case, the agent has to maximize its expected return over an indefinite period, possibly infinite. However, it can be desirable to still use time limits in order to diversify the agent's experience. For example, starting from highly diverse states can avoid converging to suboptimal policies that are limited to a fraction of the state space. In Section \ref{sec:time-unlimited}, we show that in order to learn good policies that continue beyond the time limit, it is important to differentiate between the terminations that are due to time limits and those from the environment. Specifically, for bootstrapping methods, we argue to bootstrap at states where termination is due to time limits, or more generally any other causes than the environmental ones. We refer to this approach as \textit{partial-episode bootstrapping} (PEB) and demonstrate that it can significantly improve the performance of agents.

We evaluate the impact of these considerations on a range of novel and popular benchmark domains using tabular Q-learning and Proximal Policy Optimization (PPO), a modern deep reinforcement learning \citep{Arulkumaran:2017drlbrief, Henderson:2017matters} algorithm which has recently been used to achieve state-of-the-art performance in many domains \citep{Schulman:2017ppo, Heess:2017emergence}. We use the OpenAI Baselines \citep{baselines} implementation of PPO with the hyperparameters reported by \citet{Schulman:2017ppo}, unless stated otherwise. The time-aware version of PPO concatenates the observations provided by the environment and the remaining time represented by a scalar (normalized from $-1$ to $1$). The partial-episode bootstrapping version of PPO makes a distinction between environment resets and terminations by using the value of the last state in the evaluation of the advantages if no termination is encountered. All novel tasks are implemented using OpenAI Gym \citep{Brockman:2016gym} and the standard benchmarks are from the MuJoCo \citep{Todorov:2012mujoco} Gym collection. For each task involving PPO, to achieve perfect reproducibility, we used the same $10$ seeds ($0, 1000, ..., 9000$) to initialize the pseudo-random number generators for the agents and environments.

We empirically show that time-awareness significantly improves the performance of PPO for the time-limited tasks and can sometimes result in interesting behaviors. For example, in the Hopper-v1 domain, our agent learns to efficiently jump forward and fall towards the end of its time in order to maximize its travelled distance, performing a ``photo finish''. For the time-unlimited tasks, we show that bootstrapping at the end of partial episodes allows to significantly outperform the standard PPO. In particular, on Hopper-v1, even if trained with episodes of only $200$ steps, the agent with partial-episode bootstrapping manages to learn to hop for at least $10^6$ time steps (two hours). Finally, we demonstrate that the negative impact of large experience replay buffers shown by \citet{Zhang:2017deeper} can often be vastly reduced if timeout terminations are properly handled. The source code and videos can be found at: 
\begingroup
    \fontsize{9.75pt}{12pt}\selectfont
    \mbox{\url{sites.google.com/view/time-limits-in-rl}}.
\endgroup

While the importance of time-awareness for optimizing a time-limited objective (finite horizon) is well-established in the dynamic programming and optimal control literature \citep{Bertsekas:1995dynamic, Bertsekas:1996neuro} (e.g. model-based backward induction), we observed that it has been largely overlooked in the reinforcement learning literature and in the design of the popular benchmarks. In the view of the above, this paper may serve as an introduction to this concept, and as the first attempt to bring it to bear on the problems and practices of reinforcement learning. The main contributions of this paper are: the thorough analysis of the specific issues which can be caused by the lack of time-awareness and the study of the impact of the discount factor in time-limited tasks, the formalization of the partial-episode bootstrapping method, and the extensive empirical evaluations demonstrating improved performance and stability of two existing reinforcement learning algorithms.

\section{Time-awareness for time-limited tasks}
\label{sec:time-aware}

In tasks that are time-limited by nature, the learning objective is to optimize the expectation of the return $G_{t:T}$ from Equation~\ref{equ:time-limited-return}. Interactions are systematically terminated at a predetermined time step $T$ if no environmental termination occurs earlier. This time-wise termination can be seen as transitioning to a terminal state whenever the time limit is reached. The states of the agent's environment, formally a \textit{Markov decision process} (MDP) \citep{Puterman:2014markov}, thus contain a notion of the remaining time used by its transition function. This time-dependent MDP can be thought of as a stack of $T$ time-independent MDPs followed by one that only transitions to a terminal state. Thus, at each time step $t \in \{0,...,T-1\}$, actions result in transitioning to a next state in the next MDP in the stack.

In effect, a time-unaware agent has to act in a \textit{partially observable Markov decision process} (POMDP) \citep{Lovejoy:1991POMDPsurvey} where states that only differ by their remaining time appear identical. This phenomenon is a form of \textit{state aliasing} \citep{Whitehead:1991aliasing} that is known to lead to suboptimal policies and instability due to the infeasibility of correct \textit{credit assignment}. In this case, the terminations due to time limits can only be interpreted as part of the environment's stochasticity where the time-unaware agent perceives a chance of transitioning to a terminal state from any given state. In fact, this perceived stochasticity depends on the agent's current behavioral policy. For example, an agent could choose to stay in a fixed initial state during the entire course of an episode and perceive the probability of termination from that state to be $1/T$, whereas it could choose to always move away from it in which case this probability would be perceived as zero.

In the view of the above, we consider time-awareness for reinforcement learning agents in time-limited domains by including directly the remaining time $T-t$ in the agent's representation of the environment's state or by providing a way to infer it. The importance of the inclusion of a notion of time in time-limited problems was first demonstrated in the reinforcement learning literature by \citet{harada:1997time}, yet seems to have been largely overlooked. A major difference between the approach of \citet{harada:1997time} (i.e. the Q\textsubscript{T}-learning algorithm) and that described in this paper, however, is that we consider a more general class of time-dependent MDPs where the reward distribution and the transitions can also be time-dependent, preventing the possibility to consider multiple time instances at once.

Here, we illustrate the issues faced by time-unaware agents via exemplifying the case for value-based methods. The state-value function at time $t$ for a time-aware agent in an environment with time limit $T$ is:
\begin{equation}
v_\pi(s,T-t) = \E_\pi \left[ G_{t:T} ~| ~S_t = s \right]
\label{equ:state-value-ta}
\end{equation}
By denoting an estimate of the state-value function by $\hat{v}_\pi$, the target $y$ for a one-step \textit{temporal-difference} (TD) update \citep{Sutton:1988TD}, after transitioning to a state $s'$ and receiving a reward $r$, is:
\begin{equation}
y = 
\begin{cases}
r                                                      & \quad \text{at all terminations} \\
r + \gamma \hat{v}_\pi(s',T-t-1) & \quad \text{otherwise}
\end{cases}
\label{equ:td-targets-ta}
\end{equation}
A time-unaware agent, deprived of the remaining time information, would learn value functions with or without bootstrapping from the estimated value of $s'$ depending on whether the time limit is reached. These conflicting updates for estimating the value of the same state result in an inaccurate average. It is worth noting that, for time-aware agents, if the time limit is never varied, the inclusion of the elapsed time $t$ would be sufficient. This could then be measured by the agent itself from the beginning of the current episode. For more generality, however, we chose to always represent the remaining time.

\subsection{The Last Moment problem}

\begin{wrapfigure}{r}{.17\textwidth}
    \vspace{-1.64em}
    \begin{center}
        \includegraphics[width=\linewidth]{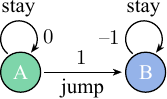}
    \end{center}
    \vspace{-1.5em}
\end{wrapfigure}
To give a simple example of the learning of an optimal time-dependent policy, we consider an MDP containing two states A and B. The agent always starts in state A and has the possibility to choose an action to ``stay'' in place with no rewards or a ``jump'' action that transitions it to state B with a reward of $1$. However, state B is a trap with no exit where the only possible action provides a penalty of $-1$. The episodes terminate after a fixed number of steps $T$. The goal of the game is thus to jump just before the timeout. For a time-unaware agent and $T > 1$, the task is impossible to master. The optimal stochastic strategy being to jump $50\%$ of the time when $T = 2$ and to never jump when $T > 2$. In contrast, a time-aware agent can learn to stay in place for $T - 1$ steps and then jump.

\subsection{The Two-Goal Gridworld problem}
\label{sec:time-aware-gridworld}

To further illustrate the impact of state aliasing for time-unaware agents, we consider a deterministic gridworld environment (see Figure~\ref{subfig:gridworld}) with two possible goals rewarding $50$ for reaching the top-right and $20$ for the bottom-left cells. The agent has $5$ actions: to move in cardinal directions or to stay in place. Any movement incurs a penalty of $-1$ while staying in place generates no reward. Episodes terminate after $3$ time steps or if the agent has reached a goal. The initial state is randomly selected for every episode, excluding goals. We used tabular Q-learning \citep{Watkins:1992Q} with random actions, trained until convergence with a decaying learning rate and a discount factor of $0.99$.

The time-aware agent has a state-action value table for each time step and easily learns the optimal policy which is to go for the closest goal when there is enough time, and to stay in place otherwise. For the time-unaware agent, the greedy values of the cells adjacent to the top-right and bottom-left goals converge to $49$ and $19$, respectively. Then, since $T=3$, from each remaining cell, the agent has between $1$ and $3$ steps. If it moves it receives a penalty and for $2/3$ of the times bootstraps from the successor cell. Thus, for $v(s)=\max_a q(s,a)$ and $N(s)$ denoting the neighbors of $s$, for states nonadjacent to the goals we have: $v(s) = 2/3 (-1 + \gamma \max_{s' \in N(s)} v(s')) + 1/3 (-1)$. This learned value function leads to a policy that always tries to go for the closest goal even if there is not enough time. While the final optimal policy does not actually require time information, this example clearly shows that the confusion during training due to state aliasing can create a leakage of the values to states that are out of reach. It is worth noting that, Monte Carlo methods such as REINFORCE \citep{Williams:1992simple, Sutton:2000pg} are not susceptible to this leakage as they use complete returns instead of bootstrapping. However, without awareness of the remaining time, Monte Carlo methods would still not be able to learn an optimal policy in many cases, such as the Last Moment problem.

\subsection{The Queue of Cars problem}

\begin{wrapfigure}{r}{.3\textwidth}
    \vspace{-1.6em}
    \begin{center}
        \includegraphics[width=\linewidth]{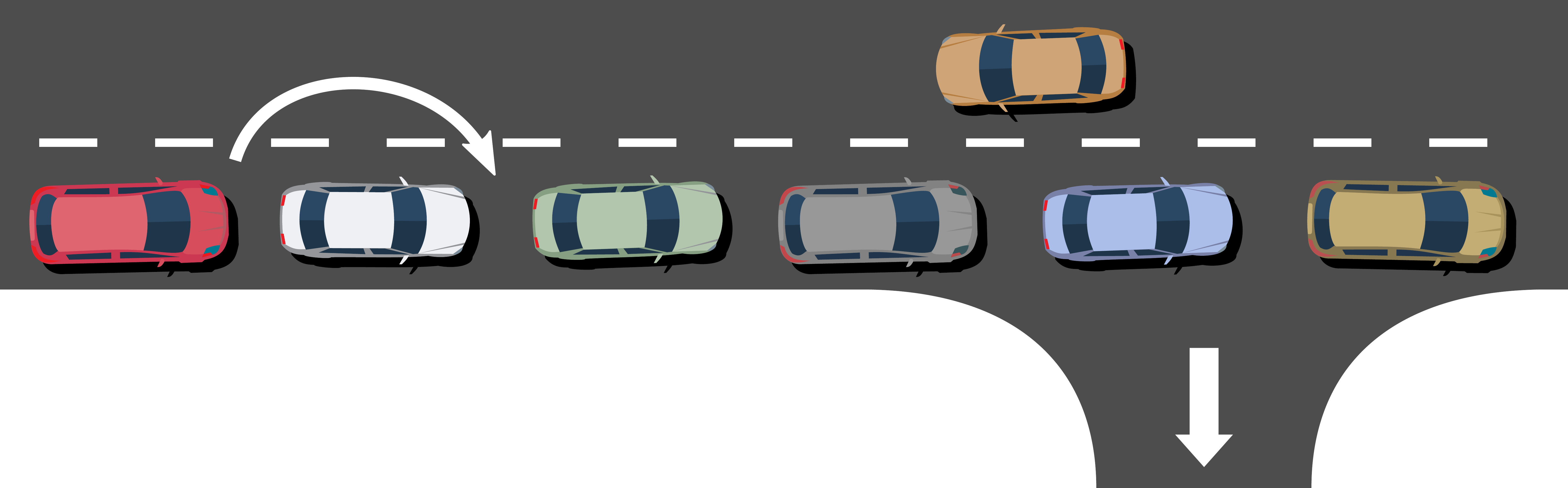}
    \end{center}
    \vspace{-1.0em}
\end{wrapfigure}

An interesting property of time-aware agents is the ability to dynamically adapt to the remaining time. To illustrate this, we introduce an environment which we call Queue of Cars where the agent controls a vehicle that is held up behind an intermittently moving queue of cars. The agent's goal is to reach an exit located $9$ slots away from its starting position. At any time, the agent can choose the ``safe'' action to stay in the queue which may result in advancing to the next slot with $50$\% probability, or to attempt to overtake with the ``dangerous'' action which has $80$\% probability to advance but poses a $10$\% chance of collision with the oncoming traffic and terminating the episode. The agent only receives a reward of $1$ when it reaches the terminal destination.

\begin{figure}[t]
    \centering
    \includegraphics[width=70mm]{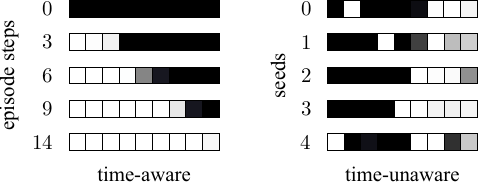}
    \caption{Heat map of the learned ``dangerous'' action probabilities overlaid on our Queue of Cars problem ($0$ and $1$ denote black and white respectively). The $9$ non-terminal states are represented from left to right. The time-aware PPO agent learns to optimally choose actions with decreasing remaining time, while standard PPO learns various suboptimal strategies depending on the initialization seeds.} \label{fig:sec1_cars_probas}
\end{figure}

Time-unaware agents cannot possibly adapt to the remaining time and thus can only learn a fixed combination of dangerous and safe actions based on the position. Figure~\ref{fig:sec1_cars_probas} shows that a time-aware PPO agent can optimally adapt to the remaining time and its distance to the goal.

\subsection{Standard control tasks}
\label{subsec:sec1_stan_control}

\begin{figure*}[t]
    \centering
    \begin{subfigure}{\textwidth}
        \centering
        \includegraphics[width=\textwidth]{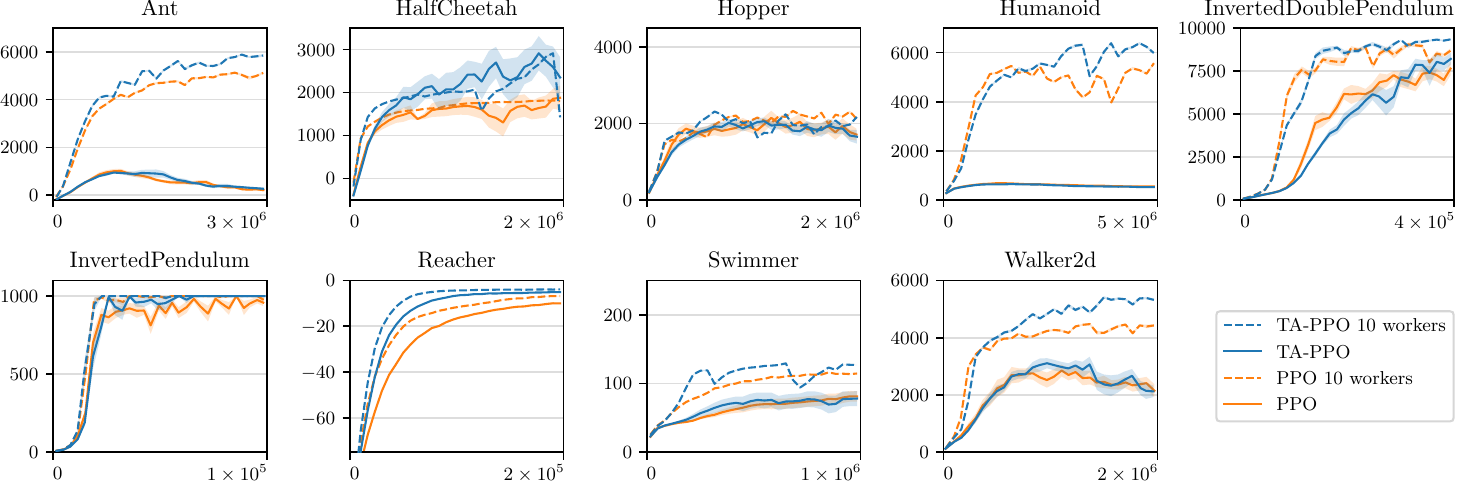}
        \caption{$\gamma=0.99$}
        \label{subfig:sec1_control_tasks_gamma099}
    \end{subfigure} \\ \vspace{2mm}
    \begin{subfigure}{\textwidth}
        \centering
        \includegraphics[width=\textwidth]{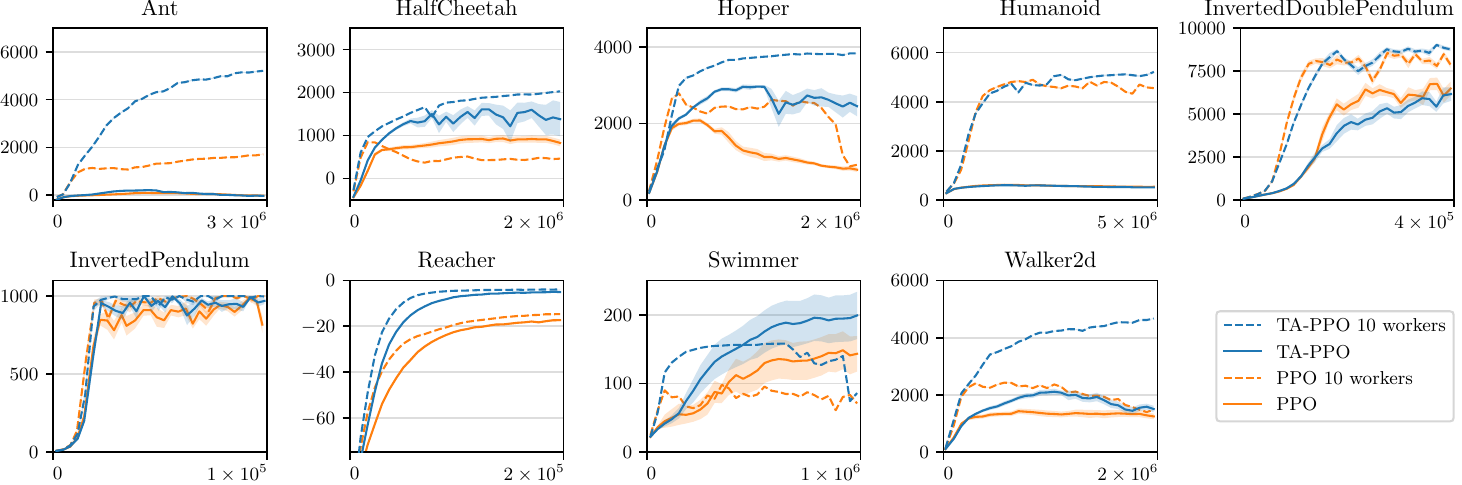}
        \caption{$\gamma=1$}
        \label{subfig:sec1_control_tasks_gamma1}
    \end{subfigure}
    \caption{Performance comparison of PPO with and without the remaining time in input on several continuous control tasks from OpenAI Gym ($T=1000$ for all except $T=50$ for Reacher-v1). The averaged sum of rewards and standard errors are shown with respect to the number of training steps. On multiple tasks, the proposed time-aware PPO (TA-PPO) outperforms the standard PPO, especially for the case with a large discount factor.}
    \label{fig:sec1_control_tasks}
\end{figure*}

In this section, we compare the performance of PPO with and without the remaining time as part of the agent's input on $9$ continuous control tasks from OpenAI Gym \citep{Brockman:2016gym, Duan:2016benchmarking}. By default, these environments use time limits which are perceived as environmental terminations. The results in Figure~\ref{fig:sec1_control_tasks} demonstrate that time-awareness (TA) significantly improves the performance and stability of PPO. To better understand the differences between the agents we now provide several more observations.

As illustrated in Figure~\ref{subfig:sec1_control_tasks_gamma099}, for a discount rate of $0.99$, often, the standard PPO is initially on par with the time-aware PPO and later starts to plateau (e.g. Walker2d-v1 and Humanoid-v1). This is due to the fact that, in some domains, the agents start to experience terminations due to the time limit more frequently as they become better, at which point the time-unaware agent begins to perceive inconsistent returns for seemingly similar states. The advantage of time-awareness becomes even clearer in the case of a discount rate of $1$ where the time-unaware PPO often diverges drastically (see Figure~\ref{subfig:sec1_control_tasks_gamma1}). This is mainly because, in this case, the time-unaware agent experiences much more significant conflicts as returns are now the sum of the undiscounted rewards.

\begin{figure}[t]
    \centering
    \includegraphics[width=\linewidth]{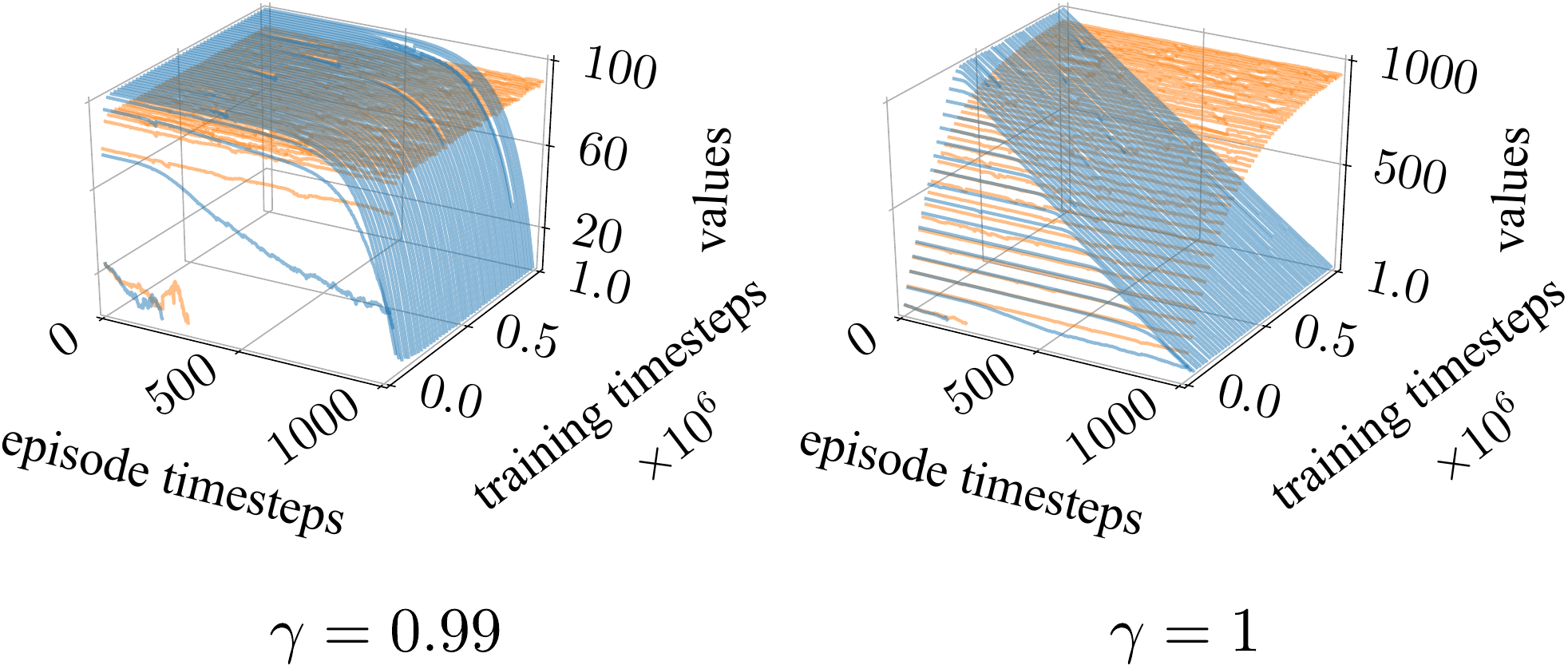}
    \label{subfig:sec1_invpend}
    \caption{
    Learned state-value estimations on InvertedPendulum-v1 ($T=1000$) of the time-aware (blue) and the standard (orange) PPO agents. The first one quickly learns a function that accurately takes into account the remaining time, while the second one slowly learns an average.
    }
    \label{fig:sec1_inverted_statevalues}
\end{figure}

Figure~\ref{fig:sec1_inverted_statevalues} shows the learned state-value estimations for InvertedPendulum-v1 which perfectly illustrate the difference between a time-aware agent and a time-unaware one in terms of their estimated expected returns. While time-awareness enables PPO to learn an accurate exponential or linear decay of the expected return with time, the time-unaware one only learns a constant estimate.

\begin{figure}[t]
    \centering
    \includegraphics[width=.7\linewidth]{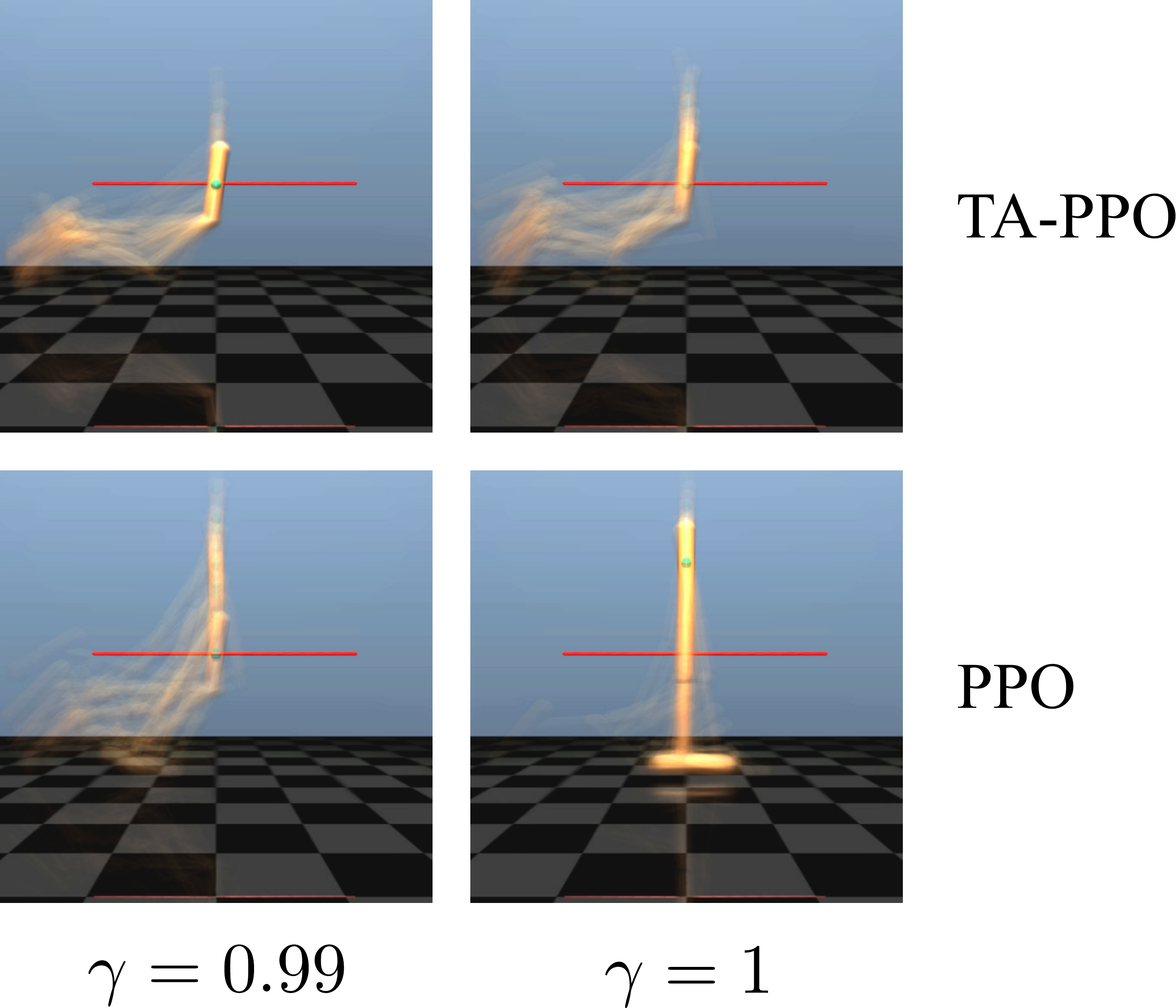}
    \caption{
    Average last pose on Hopper-v1 ($T=~300$) with the vertical termination threshold of $0.7$ meters in red. The time-aware agent (TA-PPO) learns to jump forward just before the time limit in order to maximize its forward distance. The time-unaware PPO agent does not learn this behavior and its training is highly destabilized when the discount factor is large.
    }
    \label{fig:sec1_hopper_photofin}
\end{figure}

Time-awareness does not only help agents by avoiding the conflicting updates. In fact, in naturally time-limited tasks where the agents have to maximize their performance for a limited time, time-aware agents can demonstrate interesting ways of achieving this objective. Figure~\ref{fig:sec1_hopper_photofin} shows the average final pose of the time-aware (top) and time-unaware (bottom) agents. We can see that the time-aware agent robustly learns to jump towards the end of its time in order to maximize its expected return, resulting in a ``photo finish''. It is also interesting to notice that for $\gamma = 1$, the time-unaware PPO (bottom-right) learns to actively stay in place to at least accumulate the rewards for staying alive.

In this section, we explored the scenario where the aim is to learn a policy that maximizes the expected return over a limited time. We argued for the inclusion of a notion of the remaining time as part of the agent's observation to avoid state aliasing which can cause suboptimal policies and instability. However, this scenario is not always ideal as there are cases where, even though the agent experiences time limits in its interaction with the environment, the objective is to learn a policy for a time-unlimited task. For instance, as we saw for Hopper-v1, the learned policy that maximizes the return over $300$ steps generally results in a photo finish which would lead to a fall and subsequent termination if the simulation was to be extended. Such a policy is not viable if the goal is to learn to move forward for an indefinite period. While one solution could be to not use time limits during training, short episodes remain useful to diversify experiences. In the next section, we therefore explore how time-unlimited tasks can be solved with partial episodes.

\section{Partial-episode bootstrapping for time-unlimited tasks}
\label{sec:time-unlimited}

In tasks that are not time-limited by nature, the learning objective is to optimize the expectation of the return $G_t$ from Equation~\ref{equ:time-unlimited-return}. While the agent has to maximize its expected return over an indefinite (possibly infinite) period, it is desirable to still use time limits in order to frequently reset the environment and increase the diversity of the agent's experiences. A common mistake, however, is to then consider the terminations due to such time limits as environmental ones. This is equivalent to optimizing for returns $G_{t:T}$ (Equation~\ref{equ:time-limited-return}), not accounting for the possible future rewards that could have been experienced if no time limits were used.

In the case of bootstrapping methods, we argue for continuing to bootstrap at states where termination is due to the time limit. The state-value function of a policy at time $t$ can be rewritten in terms of the time-limited return $G_{t:T}$ and the value from the last state $v_\pi(S_T)$:
\begin{equation}
v_\pi(s) = \E_\pi \left[ G_{t:T} + \gamma^{T-t} v_\pi(S_T) ~| ~S_t = s \right]
\end{equation}
By denoting an estimate of the state-value function by $\hat{v}_\pi$, the target $y$ for a one-step TD update, after transitioning to a state $s'$ and receiving a reward $r$, is:
\begin{equation}
y = 
\begin{cases}
r                          & \text{at environmental terminations} \\
r + \gamma \hat{v}_\pi(s') & \text{otherwise (including timeouts)}
    \label{equ:time-unlimited-boots}
\end{cases}
\end{equation}
An agent without partial-episode bootstrapping would not bootstrap at timeout terminations. Similarly to Equation \ref{equ:td-targets-ta}, the conflicting updates for estimating the value of the same state lead to an approximate average of these updates.

In the previous section, one of the issues came from bootstrapping values from states that were out-of-reach, letting the agent falsely believe that more rewards were available after. On the opposite, the problem presented here is when systematic bootstrapping is not performed from states at the time limit and thus, forgetting that more rewards would actually be available thereafter.

Related to the proposed partial-episode bootstrapping (PEB), \citet{white2016unifying} introduces a way to consider episodic tasks as a continuing one with a variable discount factor between them. Our approach however differs in several ways: (1)~ PEB is more suitable for tasks that do not necessarily have an underlying episodic structure such as Hopper-v1. (2) In the proposed PEB approach, the agent does not experience a transition from the last state of a partial episode to the first state of the next episode, thus enabling environmental resets based on time limits during training. (3) PEB uses a constant discount factor, allowing the learning of correct indefinite-horizon policies from partial episodes.

\begin{figure*}[t]
    \centering
    \includegraphics[width=\textwidth]{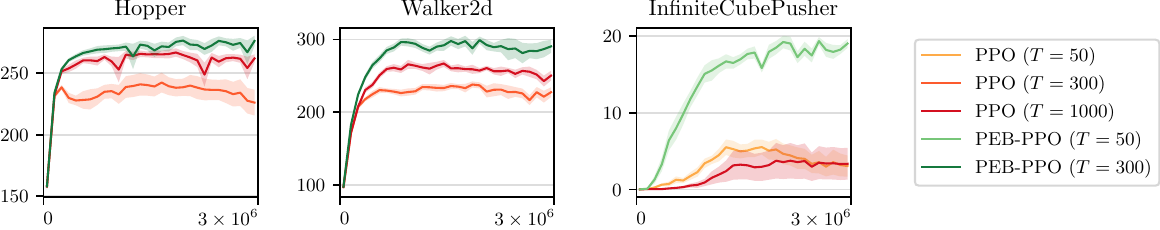}
    \caption{Performance comparison of PPO with and without partial-episode bootstrapping with $\gamma=0.99$ on several continuous control tasks. The averaged scores and standard errors are shown with respect to the number of training steps. For Hopper-v1 and Walker2d-v1, the evaluation episodes are limited to $10^6$ time steps and the discounted sum of rewards is represented, while for InfiniteCubePusher-v0 the evaluations are limited to $1000$ time steps and the number of targets reached per episode is represented.}
    \label{fig:sec2_control}
\end{figure*}

\subsection{The Two-Goals Gridworld problem}

We revisit the gridworld environment from Section~\ref{sec:time-aware-gridworld}. While previously the agent's task was to learn an optimal policy for a given time limit, we now consider how an agent can learn a good policy for an indefinite period from partial-episode experiences. The same setup and tabular Q-learning from Section~\ref{sec:time-aware-gridworld} were used, but instead of considering terminations due to time limits as environmental ones, bootstrapping is maintained from the non-terminal states that are reached at the time limits. This modification allows our agent to learn the time-unlimited optimal policy of always going for the most rewarding goal (see Figure~\ref{subfig:gridworld_peb}). On the other hand, while the standard agent that is not performing the final-step bootstrapping (see Figure \ref{subfig:gridworld_std}) has values from out-of-reach cells leaking into its learned value function, these updates do not occur in sufficient proportion to let the agent learn the time-unlimited optimal policy.

For the next experiments, we again used PPO but with two key modifications for partial-episode-bootstrapping. First, we removed the Gym's TimeLimit wrapper that is included by default for all environments and which enforces termination when time limits are reached. Second, we modified the PPO's implementation to enable continuing to bootstrap when the environment is reset but no termination is encountered. This involves changing the implementation of the generalized advantage estimator (GAE) \citep{Schulman:2016gae}. Whereas GAE uses an exponentially-weighted average of $n$-step value estimations for bootstrapping which is more complex than the one-step lookahead bootstrapping explained in Equation~\ref{equ:time-unlimited-boots}, continuing to bootstrap from the last non-terminal states is the only modification required for the considered approach.

\subsection{Hopper and Walker}

Here, we consider the Hopper-v1 and Walker2d-v1 environments from Section~\ref{subsec:sec1_stan_control}, but instead aim to learn a policy that maximizes the agent's expected return over a time-unlimited horizon. The goal here is to show that by continuing to bootstrap from states at timeout terminations it is possible to learn good policies for time-unlimited domains. Figure~\ref{fig:sec2_control} (left and middle) demonstrates performance evaluations of the standard PPO against one with partial-episode bootstrapping (PEB). During training, partial episodes were limited to $300$ time steps, while during evaluation episodes were limited to $10^6$ time steps to prevent infinitely long evaluation episodes. The standard PPO agent was also trained with the default time limit ($T=1000$) for comparison. The results show that partial-episode bootstrapping allows our agent to significantly outperform the standard PPO and that training on short interactions is sufficient.

\subsection{The Infinite Cube Pusher task}

\begin{wrapfigure}{r}{3.3cm}
    \vspace{-1.2em}
    \includegraphics[width=3.3cm]{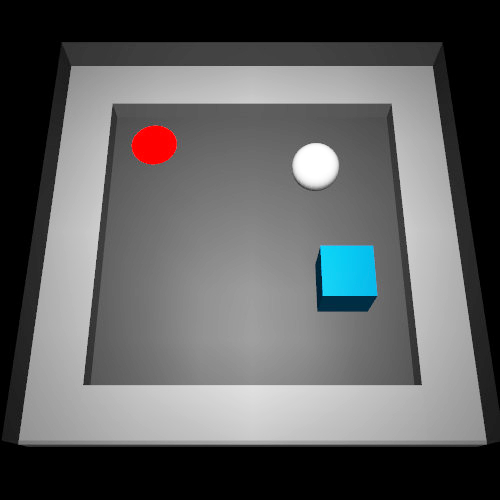}
    \vspace{-1.4em}
\end{wrapfigure}
To demonstrate the ability of our agent in optimizing for an infinite-horizon (no terminal state) objective, we propose a novel MuJoCo domain consisting of a torque-controlled ball, on the horizontal plane, that is used to push a cube to specified target positions. Once the cube has touched the target, the agent is rewarded and the target is moved away from the cube to a new random position. Because the task lacks terminal states, it can continue indefinitely. The terrain is surrounded by fixed bounding walls. The inner edge of the walls stops the cube but not the ball in order to let the agent move the cube even if it is in a corner. The environment's state representation consists of the coordinates of the ball, the cube, and the target, the velocities of the ball and the cube, and the rotation of the cube. The agent receives a reward of $1$ every time the cube reaches the target. Due to the absence of \textit{reward shaping} \cite{ng1999policy}, it is necessary to limit training episodes in time to diversify the experiences and learn to solve the task. Therefore, during training, a time limit of $50$ time steps was used, sufficient to push the cube to one target in most cases. During evaluation, however, $1000$ steps were used to allow successfully reaching several targets. An entropy coefficient of $0.01$ was used to encourage exploration. We found this value to yield best performance for both agents. Figure~\ref{fig:sec2_control} (right) shows that our agent drastically outperforms the standard PPO.

\begin{figure*}[t]
    \includegraphics[height=3.4cm]{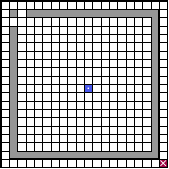}
    \hfill
    \includegraphics[height=3.4cm]{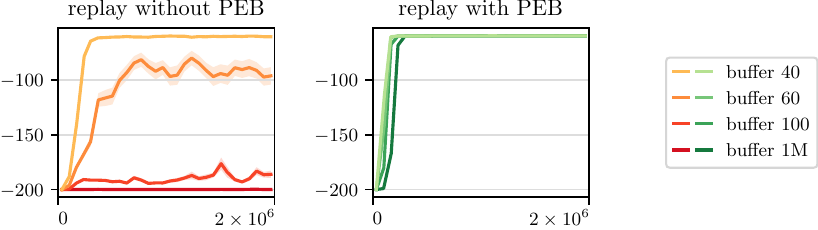}
    \caption{Performance comparison of tabular Q-learning with and without partial-episode bootstrapping with $\gamma = 1$ on the Difficult Gridworld ($T = 200$) problem presented in \cite{Zhang:2017deeper}. The averaged scores and standard errors are shown with respect to the number of training steps. When timeout terminations are not properly considered, experience replay significantly hurts the performance, while by simply continuing to bootstrap whenever a timeout termination is encountered, the learning is much faster and varying the buffer size has almost no effect.}
    \label{fig:sec2_experience_replay}
\end{figure*}

\subsection{Experience replay}

Sampling batches of transitions from a buffer of past experience, known as \textit{experience replay} \cite{lin1992self}, has proved to be highly effective in stabilizing the training of artificial neural networks by decorrelating updates and avoiding the rapid forgetting of rare experiences \cite{Mnih:2015natureDQN, Schaul:2016prior}. However, we argue that the perceived non-stationarity, induced by not properly handling time limits, is incompatible with experience replay. Indeed, the timeout-occurrence distribution changes with the behavior of the agent, and thus past transitions become obsolete.

While both time-awareness and partial-episode bootstrapping (PEB) provide ways to solve this issue, we chose to illustrate the effect of PEB on one of the tasks presented in \cite{Zhang:2017deeper}. In the latter, the authors demonstrate that experience replay can significantly hurt the learning process if the size of the replay buffer is not tuned well. One of the environments used is a deterministic gridworld with a fixed starting state and goal, shown in Figure~\ref{fig:sec2_experience_replay}. As proposed by the authors, tabular Q-learning is used with values initialized to $0$, a penalty of $-1$ at each time step, no discount, a time limit $T=200$, and an $\varepsilon$-greedy exploration using a fixed $10\%$ chance of random actions. Figure \ref{fig:sec2_experience_replay} shows the performance with respect to the number of training steps, averaged over $30$ seeds, from $0$ to $29$. We successfully replicated the figure showing that the performance deteriorates very quickly with buffer size and demonstrate that by simply bootstrapping from states when the time limit is reached the effect of the buffer size is vastly diminished.

\section{Discussion}
\label{sec:discussion}

We showed in Section~\ref{sec:time-aware} that time-awareness is required for correct credit assignment in domains where the agent has to optimize its performance over a time-limited horizon. However, common time-unaware agents still often manage to perform relatively well. This could be due to several reasons including: if time limits are so long that timeouts are hardly ever experienced (e.g. in the Arcade Learning Environment (ALE) \citep{Bellemare:2013arcade, Machado:2017revisitALE} domains where $T=5$ minutes), if there are clues in the observations that are correlated with time (e.g. the forward distance), if it is not likely to observe the same states at different remaining times, or if the discount factor is sufficiently small to reduce the impact of the confusion. Furthermore, many methods exist to handle POMDPs \citep{Lovejoy:1991POMDPsurvey}. In deep learning \citep{Lecun:2015deep, Schmidhuber:2015NN}, it is highly common to use a stack of previous observations or recurrent neural networks (RNNs) \citep{Goodfellow:2016DLbook} to address partial observations \citep{Wierstra:2009rpg}. These solutions may to an extent help when a notion of the remaining time is not included as part of the agent's input. However, including this information is much simpler and allows better diagnosis of the learned policies. The considered approach is rather generic and can be applied to domains with varying time limits. It is also interesting to note that time-awareness can allow, to an extent, the agent to learn open-loop (state-independent) policies which can be easier to learn than closed-loop ones. For example, if a task involves a fixed starting state and no stochastic transitions, then an optimal policy can rely only on the time. Finally, in real-world applications, such as robotics, the real clock time can be used in place of discrete fixed time steps.

In order for the partial-episode bootstrapping method in Section~\ref{sec:time-unlimited} to work, as is the case for bootstrapping methods in general, the agent needs to use reliable predictions. This is in general resolved by enabling sufficient exploration. However, when the interactions are limited in time, exploration of the full state-space may not be feasible from fixed starting states. Thus, a good way to allow appropriate exploration in such domains is to sufficiently randomize the initial states. It is worth noting that partial-episode bootstrapping is generic in that it is not restricted to partial episodes only due to time limits. In fact, this approach is valid for any early termination causes. For example, it is common in curriculum learning to start from states nearby the goals and gradually expand to further ones \citep{Florensa:2017reverse}. In this case, it can be helpful to stitch the learned values by terminating the episodes and bootstrapping as soon as the agent enters a well-known state.

\section{Conclusion}
\label{sec:conclusions}

We considered the problem of learning optimal policies in time-limited and time-unlimited domains using time-limited interactions. We showed that time limits should be carefully manipulated to avoid state aliasing and perceived non-stationarity of the environment. We explained that when learning policies for time-limited tasks, it is important to include a notion of the remaining time as part of the agent's input. We then showed that, when learning policies for time-unlimited tasks, it is necessary for correct value estimation, to continue bootstrapping at the end of the partial episodes when termination is due to time limits, or any early termination causes other than the environmental ones. In both cases we observed significant improvements in the performance of the considered reinforcement learning algorithms.

\section*{Acknowledgments}

The research presented in this paper has been supported by Dyson Technology Ltd., Samsung, and EPSRC. We also acknowledge computation resources provided by Microsoft via a Microsoft Azure award.

\bibliographystyle{icml2018}
\bibliography{refs}

\end{document}